\PassOptionsToPackage{unicode}{hyperref}
\documentclass[review]{elsarticle}
\usepackage{lineno,hyperref}
\usepackage{amsmath}
\usepackage{graphicx}
\usepackage[linesnumbered,algoruled,boxed,lined]{algorithm2e}
\usepackage[titletoc]{appendix}

\journal{Computers \& Operations Research}

\makeatletter 
\g@addto@macro{\@algocf@init}{\SetKwInOut{Parameter}{Parameters}} 
\makeatother
\modulolinenumbers[5]
\biboptions{authoryear}

\usepackage{amsmath}

\pdfoutput=1
\begin{document}
\nolinenumbers
\begin{frontmatter}

\title{Autoselection of the Ensemble of Convolutional Neural Networks with Second-Order Cone Programming}


\author[mymainaddress]{Buse~\c{C}isil~G\"{u}ldo\u{g}u\c{s} \corref{mycorrespondingauthor}}
\cortext[mycorrespondingauthor]{Corresponding author}
\ead{busecisi.guldogus@bahcesehir.edu.tr}

\author[mysecondaryaddress]{Abdullah~Nazhat~Abdullah}
\ead{nazhat.abdullah@bahcesehir.edu.tr}

\author[mysecondaryaddress]{Muhammad~Ammar~Ali} 
\ead{muhammadammar.ali@bahcesehir.edu.tr} 

\author[mythirdaddress]{S\"{u}reyya~\"{O}z\"{o}\u{g}\"{u}r-Aky\"{u}z}
\ead{sureyya.akyuz@eng.bau.edu.tr}

\address[mymainaddress]{Department of Industrial Engineering, Graduate School of Natural and Applied Sciences, Bahcesehir University, Istanbul, Turkey}
\address[mysecondaryaddress]{Department of Computer Engineering, Graduate School of Natural and Applied Sciences, Bahcesehir University, Istanbul, Turkey}
\address[mythirdaddress]{Department of Mathematics, Faculty of Engineering and Natural Sciences, Bahcesehir University, Istanbul, Turkey}



\begin{abstract}
Ensemble techniques are frequently encountered in machine learning and engineering problems since the method combines different models and produces an optimal predictive solution. The ensemble concept can be adapted to deep learning models to provide robustness and reliability. Due to the growth of the models in deep learning, using ensemble pruning is highly important to deal with computational complexity. Hence, this study proposes a mathematical model which prunes the ensemble of Convolutional Neural Networks (CNN) consisting of different depths and layers that maximizes accuracy and diversity simultaneously with a sparse second order conic optimization model. The proposed model is tested on CIFAR-10, CIFAR-100 and MNIST data sets which gives promising results while reducing the complexity of models, significantly.
\end{abstract}


\end{frontmatter}

\nolinenumbers

\section{Introduction}
Machine learning has made great progress in the last few years due to advances in the use of Deep Neural Networks (DNN), but many of the proposed neural architectures come with high computational and memory requirements \cite{Blalock20}. These demands increase the cost of training and deploying these architectures, and constrain the spectrum of devices that they can be used on.

Although deep learning has been very promising in many areas in terms of technology in recent years, it has some problems that need improvement. Even though deep learning can distinguish changes and subtle differences in data with interconnected neural networks, it makes it very difficult to define hyperparameters and determine their values before training the data. For this reason, different pruning methods have been proposed in the literature to reduce the parameters of convolutional networks \cite{Han16,Hanson88,lecun-mnisthandwrittendigit-2010,Strom97}.

The common problem of deep learning and pruning algorithms that have been studied in recent years is the decision of the pruning percentage with the heuristic approach at the pruning stage, making the success percentage of the deep learning algorithm dependent on the pruning parameter. On the other hand, the optimization models proposed with zero-norm penalty to ensure sparsity ignore the diversity of the layers as they only take into account the percentage of success. The combination of the layers that are close to each other does not increase the percentage of accuracy.

As the primal example of DNNs, Convolutional Neural Networks (CNNs) are feed-forward architectures originally proposed to perform image processing tasks, \cite{CNN_Survey} but they offered such high versatility and capacity that allowed them to be used in many other tasks including time series prediction, signal identification and natural language processing.
Inspired by a biological visual perceptron that displays local receptive fields \cite{Goodfellow-et-al-2016}, a CNN uses learnable kernels to extract the relevant features at each processing level. This provides the advantage of local processing of information and weight sharing. Additionally, the use of the pooling mechanism allows the data to be down-sampled and the features to be invariant representations of previous features.

Recently, a great deal of attention has been paid to obtain and train compact CNNs with the effects of pruning \cite{Wen16,Zhou16}. Generally, pruning techniques differ from each other in the choice of structure, evaluation, scheduling, and fine-tuning \cite{Blalock20}. However, tuning these choices according to its costs and requirements, the network should be pruned. This is a systematic reduction in neural network parameters while aiming to maintain a performance profile comparable to the original architecture \cite{Blalock20}. Pruning of parameters in neural networks has been used to reduce the complexity of the models and prevent overfitting. Different methods on pruning in deep neural networks are proposed in the literature such as; Biased Weight Decay \cite{Hanson88}, Optimal Brain Damage \cite{Cun89} and Optimal Brain Surgeon \cite{Hassibi92}. These aim to identify redundant parameters represented by connections within architectures and perform pruning by removing such redundancies \cite{Srini15,Han16}. In another study, a new layer-based pruning technique was proposed for deep neural networks where the parameters of each layer are pruned independently according to the second-order derivatives of a layer-based error function with respect to the relevant parameters \cite{Dong17}. Poor estimation performance is observed after pruning is limited to a linear combination of reconstructed errors occurring in each layer \cite{Dong17}. Some pruning techniques, such as unstructured pruning,  work on individual parameters that induce sparsity while the memory footprint of the architecture reduces although there is not any speed advantage. Other techniques work on groups of parameters, where a neuron filter layer is considered as a separate target for the algorithm, thus provides the ability to take advantage of existing computational hardware \cite{Li16}. 

Evaluation of pruned parameters can be based on relevance to training metrics, absolute value, activation, and contribution of gradients \cite{Blalock20}. The architecture can be considered and evaluated as a whole \cite{Lee19,Frank19}, while parameters can be scored locally with reference to a small group of closely related groups of networks \cite{Han16}. Iterative pruning is performed by considering a subset of the architecture \cite{Han16} and the speed of the pruning algorithm varies in a single step process \cite{Liu19}. Changing the pruning rate according to a certain rule have also been tried and implemented in the literature \cite{Gale19}. After applying the pruning algorithm, some techniques continued to use the same training weights; while the reduced architecture obtained in the studies was retrained to a certain state \cite{Frank19} or initialized and restarted from the initial state \cite{Liu19}.

Deep Ensembling techniques increase reliability of DNNs as multiple diverse models are combined into a stronger model \cite{Fort2019DeepEA}. It was shown that members of the deep ensemble models provide different predictions on the same input increasing diversity \cite{LossLandscape} and providing more calibrated probabilities \cite{Simple_Scalable_probs_10.5555/3295222.3295387}. In one of the recent studies, ensemble technique is chosen to increase the complexity by training end-to-end two EfficientNet-b0 models with bagging. Adaptive ensemble technique is used by fine-tuning within a trainable combination layer which outperforms different studies for widely known datasets such as CIFAR-10 and CIFAR-100 \cite{Bruno22}. Since pre-trained models boosts efficiency while simplifying the hyperparameter tuning, increasing the performance on these datasets are achieved with the help of transfer learning and pre-training \cite{Koles20,Sun17}. Researchers accelerated CNNs but get lower performance metrics on CIFAR-100. One of these techniques is FATNET algorithm where high resolution kernels are used for classification while reducing the number of trainable parameters based on Fourier transform \cite{Ibad22}. 

Ensemble pruning methods are used to reduce the computational complexity of ensemble models, and remove the duplicate models existing in the ensemble. Finding a small subset of classifiers that perform equivalent to a boosted ensemble is an NP-hard problem \cite{tamon2000boosting}. Search based methods can be used to select an ensemble as an alternative. These methods conduct a search in the ensemble space and evaluate performance of various subsets of classifiers \cite{sagi2018ensemble}. Deep learning can actually be thought as multi-layer artificial neural networks, which is an example of ensemble learning. The overall percentage of success for ensemble learning is proportional to the percentage of average accuracy of the ensemble and the diversity of each learner within the ensemble. However, the percentage of accuracy and diversity have trade-offs among themselves. In other words, an increase in the percentage of accuracy within the community causes a decrease in diversity, which means redundant merge of similar layers. It has been shown that a pruning problem can be re-framed to a quadratic integer problem for classification to seek a subset that optimizes accuracy-diversity trade-off using a semi-definite programming technique \cite{zhang2006ensemble}. In the vein of that research, multiple optimization models have been proposed to utilize the quadratic nature of pruning problems. 

Sparse problem minimization and low-order approximation techniques have recently been used in numerous fields such as computer vision, machine learning, telecommunications, and more \cite{Quach2017NonconvexOR}. In such problems, including zero-norm corresponds to sparsity which makes the objective function non-convex in the optimization problem. For this reason, different relaxation techniques have been proposed in the optimization literature for non-convex objective functions. For class of a non-convex quadratic constrained problem, SDP relaxation techniques have been developed based on the difference of two convex functions (Difference of Convex – DC) decomposition strategy \cite{Zheng2011NonconvexQC,Akyuz20}. However, since SDP algorithms for high-dimensional problems are slow and occupy a lot of memory, SDP algorithms have been relaxed into quadratic conic problems. 

Second-Order Cone programming (SOCP) is a problem-solving class that lies between linear (LP) or quadratic programming (QP) and semi-definitive programming (SDP). Like LP and SDP, SOCPs can be solved efficiently with primal dual interior-point methods \cite{Potra00}. Also, various engineering problems can be formulated as quadratic conic problems \cite{Lobo98}. In the literature, new convex relaxations are suggested for non-convex quadratically constrained quadratic programming (QCQP) problems related to this issue. Since basic semidefinite programming relaxation is often too loose for general QCQP, recent studies have focused on enhancing convex relaxations using valid linear or SOC inequalities. Valid second-order cone constraints were created for the non-convex QCQP and the duality difference was reduced by using these valid constraints \cite{Jiang19}. In addition, a branch and bound algorithm has been developed in the literature to solve continuous optimization problems in which a non-convex objective function is minimized under non-convex inequality constraints that satisfy certain solvability assumptions \cite{Beck17}.

In this study, we propose a novel optimization model to prune the ensemble of CNNs with a Second-Order Cone Programming (SOCP) model. The proposed optimization model selects the best subset of models in the ensemble of CNNs by optimizing the accuracy-diversity trade-off while reducing the number of models and computational complexity. In the following chapters; definition of the loss function and relaxations for SOCP model will be introduced in Section \ref{materials}, developing the ensemble of CNNs with hardware, software specifications, datasets and results will be given in Section \ref{experiments} and lastly, the conclusions and future work will be mentioned in Section \ref{conclusion}, respectively. 

\subsection*{\textbf{Contribution of The Study}}
The contributions of the proposed study are listed as follows;
\begin{itemize}
\item One of the unique values that this study will add to the literature is the fact that the parameter called the pruning percentage, which will meet the needs of the deep learning literature, will be directly obtained by the proposed second-order conical optimization model. \item Another important original value will be a solution to the ignorance of the diversity criterion in the literature for pruning deep learning networks only on the basis of accuracy performance. The objective function in the proposed optimization model simultaneously optimizes the accuracy and diversity criteria by pruning the ensemble of CNNs.
\item Since SOCP models gives more successful and faster results than the other sparse optimization models. Hence, the number of models in the ensemble of CNNs will be reduced significantly by using the proposed SOCP model. 
\end{itemize}

\section{Methods and Materials} \label{materials}
Suppose that in a system consisting of $M$ binary classifiers where $i_{th}$ classifier calculates the probability that a selected sample belongs to Class $1$, as $f_i^{{1}}$. In this system, to find a consensus over all classifiers a weighted majority within the ensemble can be defined as \cite{Polikar09}:

\begin{equation}
	\label{eq1}
	f_{ens}=\sum_{i=0}^{M-1}{\omega_if_i^{{1}}}.
\end{equation}

In the Equation (\ref{eq1}), $\omega_i$ refers to weights which express the success of the classifier and these $\omega_i$ values are determined through the optimization model that will be introduced next in Equation (\ref{eq2}). If we denote the output distributions of the classifiers with $p_i$, $f_{gt}\in \{0,1\}$ where $f_{gt}$ is the ground truth, we can define the loss function of the ensemble as follows:

\begin{equation}
	\label{eq2}
	L_{ens}\left(f\right)=\alpha\left(f_{ens}-f_{gt}\right)^2+\left(1-\alpha\right)\left(1-\left(H\left(\sum_{i=0}^{M-1}{\omega_ip_i}\right)-\sum_{i=0}^{M-1}{\omega_iH(p_i)}\right)\right).
\end{equation}

In Equation (\ref{eq2}) , the first term defines the error, and the second term represents $(1-\text{diversity})$. The minimization of  $L_{ens}$ is maximization of accuracy and diversity while the trade-off between these two terms is determined by $\alpha$. The Equation (\ref{eq2}) can be rewritten for multi-class classification problems where \textit{C} is the number of classes and $f_{gt}\in \{0,1\}^C$ represents the correct classification vector:

\begin{equation}
	\begin{split}
	\label{eq3}
	L_{ens}=\alpha\sum_{j=0}^{C-1}\frac{\left(f_{ens,j}-f_{gt,j}\right)^2}{C}+ \\
	\left(1-\alpha\right)\left(1-\frac{\sum_{j=0}^{C-1}\left(H\left(\sum_{i=0}^{M-1}{\omega_ip_{i,j}}\right)-\sum_{i=0}^{M-1}{\omega_iH\left(p_{i,j}\right)}\right)}{C}\right).
\end{split}
\end{equation}
Here, the first term refers to the accuracy (quality) while the second term refers to the $(1-\text{diversity})$ defined by the Jensen Shannon Entropy function $(H(x)=-x \log x)$. The hyperparameter $\alpha$ provides the trade-off between accuracy and diversity terms. Additional information for the convex form of Shannon Entropy can be found in (\ref{shan}). Thus, ensemble pruning model which optimizes accuracy and diversity can be written as;
\begin{equation}
	\label{eq4}
	\begin{aligned}
		\min_{\omega} \quad & L(\omega;\alpha) 
	\end{aligned}
\end{equation}
where $\alpha$ is determined through cross validation. The solution of the Equation (\ref{eq4}) determines weights for each model in the ensemble. In order to obtain sparse weights, we introduced a regularization term:
\begin{equation}
	\label{eq5}
	\begin{aligned}
		\min_{\omega} \quad & L(\omega;\alpha) + \lambda \| \omega \|_0.
	\end{aligned}
\end{equation}
The Equation (\ref{eq5}) can be written in a more generalized form:
\begin{equation}
	\centering
	\min_x f(x)+\lambda \| x \|_0.
	\label{socp2}
\end{equation}
The $f(x)$ used in this Equation \ref{socp2} is a convex quadratic function as using the Shannon Entropy function above in convex form where conceptualization is given in \ref{shan}. In the Equation \ref{socp2}, $\| x \|_0$ is the zero-norm for sparsity and $\lambda$ is defined as regularization term. Since the proposed model will be applied to highly correlated and large ensemble of CNNs, a slightest change within the dataset will cause a large error which is called overfitting. In order to avoid this problem, memory usage and model size are reduced by applying regularization term $\lambda$ to the problem. The zero-norm $\| x \|_0$ is preferred to ensure sparsity and to penalize the number of non-zero entries of the coefficient vector. Here, the regularization term reflects non-zero inputs in the coefficient vector, at which quantity is usually the zero-norm of the vector. However, it is not a valid norm to use since zero-norm is not differentiable and the best convex approach for $\| x \|_0$ is $\| x \|_1$ which is convex \cite{Ramirez13}. Thus,  the  $\| x \|_0$ norm is relaxed by using the $\| x \|_1$ norm as the best approximation and the Equation \ref{socp2} can be rewritten as:

\begin{equation}
	\centering
	\min_x f(x)+\lambda \| x \|_1,
	\label{socp3}
\end{equation}
where the minimization problem becomes a convex quadratic problem. The Equation (\ref{socp3}) can be written as a SOCP problem by algebraic reformulations as follows:
\begin{equation*}
	\| x \|_1 = \sum_{i=1}^{n} |x_i| \leq u,
\end{equation*}
where $(x_i) \in K_{q}^{2}$, $i=1,\ldots,n$ and $u$ is defined to be an additional independent variable. Thus, by restricting the variable $x$ into the cone and adding the additional constraint $x \in K$, the Equation \ref{socp3} becomes:
\begin{equation*}
	\begin{aligned}
		\min \quad & f(x)+\lambda u \\
		\textrm{s.t.} \quad & \sum_{i=1}^{n} |x_i| \leq u,\\ 
		& x \in K .   \\  
	\end{aligned}
\end{equation*}
A convex cone $K$ such that $K=K^{n_1}x\ldots xK^{n_r}xR^{n_i}$ and $x_i\in K^n, K^{n_i}$ is defined as a quadratic cone with dimension $n_i$:
\begin{equation*}
	K^{n_i}=x_i=\left[\begin{matrix}x_{i_1}\\x_{i_0}\\\end{matrix}\right]\in R^{n_i-1}xR:\| x_{i_{1}} \|_2 \leq x_{i_{0}},
\end{equation*}
where $\|x_{i_{1}} \|_2$ is the standard Euclidean norm. 
Based on relaxation techniques given in \ref{quad}, an additional term $t=x^{T}Qx$ is introduced as:
\begin{equation*}
	\label{convexq1}
	\begin{aligned}
		\min \quad & t+c^{T}x\\
		\textrm{s.t.} \quad & x^{T}Qx\leq t,\\ 
	\end{aligned}
\end{equation*}
and constraints are transformed as follows: 
\begin{gather*}
		t\geq x^TQx, \\
		0\geq x^TQx-t, \\
		0\geq4x^TQx-4t, \\ 
		0\geq4x^TQx+(1-t)^2-(1+t)^2,
\end{gather*}
where $t\geq 0$. Hence, $1+t\geq0$ can be modified to the problem as an additional constraint where $t \geq 0$, $x^{T}Qx \geq 0$, $t \geq x^{T}Qx$ and since $Q$ is positive definite, Cholesky factorization $Q=LL^{T} $ can be applied \cite{Higham90}:
\begin{gather*}
1+t\geq\sqrt{4x^TQx+\left(1-t\right)^2},\ 1+t\ \geq0 \\
1+t\geq\sqrt{4x^TLL^Tx+\left(1-t\right)^2},\ 1+t\ \geq0 \\
1+t\geq2Lx1-t, 1+t\ \geq0 \\
\left[\begin{matrix}1+t\\2Lx\\1-t\\\end{matrix}\right]\ \geq 0.
\end{gather*}
Accordingly, the Equation (\ref{socp2}) can be defined as a Second-Order Cone problem:
\begin{equation}
	\label{convexq2}
	\begin{aligned}
		\min \quad & (\alpha) (t)+(1-\alpha)(c^{T}x)+\lambda u\\
		\textrm{s.t.} \quad & \left[\begin{matrix}1+t\\2Lx\\1-t\\\end{matrix}\right]\ \geq 0,\\ 
		& \sum_{i=1}^{n}\left|x_i\right|\le u,   \\
		& x \in K,
	\end{aligned}
\end{equation}
where $K=K^{n_i} \ldots K^{n_r} R^{n_i}$ when $K^{n_i}$ is either a second-order cone or a rotated second-order cone $K_r^{n_i}$ while $(x_i)  \in K_q^2.$
\section{Experiments and Results} \label{experiments}
Traditionally, the generation of ensembles uses bootstrap aggregation (bagging) techniques, however it was highlighted that random initialization of deep models in the ensemble is more effective than bagging approaches \cite{Lee2015WhyMH}. These ensembles of randomly initialized DNNs explore the function space of the architectures more extensively covering a wider span of the accuracy-diversity plane and strongly competing with methods such as weight averaging \cite{Izmailov2018AveragingWL} and local Gaussian approximations \cite{Gaussian_process_NguyenTuong2009ModelLW} that produce models with low diversity in the function space.

In this study, to achieve a statistical measure for the ensemble with the desired diversity and accuracy it was chosen to generate 300 CNN models with a process of random selection on the following design specifications:
\begin{itemize}
	\item Convolution Blocks: Each convolution block consists of two convolution layers, one batch normalization layer, ReLU (Rectified Linear Unit) non-linearity activation function and one pooling layer. The range of blocks was randomly selected from $3$ to $5$.
	\item Convolution Filters: For each convolution layer the number of convolution filters is subject to randomization with a range of $32$ to $256$ with a step size of $32$. 
	\item Pooling Mechanism : Two pooling mechanisms are randomly chosen from average pooling and maximum pooling.
	\item Classification Head (Fully Connected layer) size: The fully connected (Dense) layer has its size (number of neurons) randomized in range of $20$ to $100$ with a step of $10$.
	\item Drop Out Layer Percentage : The drop out regularization layer has its percentage randomized in range of $0$ (no Drop Out) to $0.5$ with a step of $0.1$.
	\item Learning Rate: The optimizer learning rate was randomized in a range of $1e^{-4}$ to $1e^{-2}$ with logarithmic sampling.
	\item All CNN models are trained using ADAM optimizer and Sparse Categorical Cross-Entropy loss function.
\end{itemize}
\subsection{Hardware, Software and Data Sets}
The available hardware system is a desktop personal computer with Intel i9-9900X CPU, $64$ Gigabytes of system RAM, NVIDIA RTX 2080ti GPU with $12$ Gigabytes of RAM and running an \textit{UBUNTU} $20$ LTS operating system.

The utilized software platform is \textit{Tensorflow-GPU} (ver. 2.4.1) supported by CUDA toolkit (ver. 10.1.243) and cuDNN (ver. 7.6.5). All implementations performed on \textit{Python} environment (ver. 3.9.5).

Utilized data sets are of image classification tasks where three have been chosen with differences in input size and number of classes to verify the robustness of the proposed method to changes in data distribution. Datasets chosen are as follows:
\begin{itemize}
	\item \textbf{CIFAR-10:} This data set consists of 60000 32x32 resolution colour images in 10 classes, with 6000 images per class. There are 50000 training images and 10000 test images \cite{cifar10}.
	\item \textbf{CIFAR-100:} As with CIFAR-10 the data set consists of 60000 32x32 resolution colour images, in 100 classes, resulting in 600 images per class. Similarly to CIFAR-10, There are 50000 training images and 10000 test images \cite{cifar_100}.
	\item \textbf{MNIST:} This data set contains 70000 28x28 resolution grey scale handwritten digits in 10 classes. There are 60000 training images and 10000 test images \cite{lecun-mnisthandwrittendigit-2010}.	
\end{itemize}
In the generation of the ensemble CNN models, the chosen data sets were utilized in the standard train, validation and test splits. The prediction classes and probabilities were recorded to constitute the data for the SOCP process. The samples of CIFAR-10 and CIFAR-100 is splitted as 40k for training and 10k for validation while MNIST data set is divided into 40k samples for training and 20k for validation. All test splits were comprised of 10k samples. The pseudo code of proposed ensemble pruning approach by SOCP modeling is given below in Algorithm \ref{STGY}.

\begin{algorithm}
	\scriptsize
	\KwIn{$d$: dataset, $M$: number of Convolutional Neural Network models}
	\Parameter{$b$: Convolutional Blocks, $F$: Convolutional Filters, $m$: Pooling Mechanism, $s$: Fully Connected Layer Size, $d$: Drop-Out Layer Percentage, $r$: Learning Rate, $h$: sensitivity threshold, $l$: $\Lambda$ values,  $a$: $\alpha$ values and $e$: experiment size which is $l.a$}
	\KwOut{P: Accuracy of Pruned Ensemble}  
	\BlankLine
	\For{$i\leftarrow 1$ \KwTo $M$}{
		Split $d$ for training $d_{\textit{training}}$, validation $d_{\textit{valid}}$ testing $d_{\textit{testing}}$ \\
		Randomize $b$,$F$, $m$,$s$, $d$ and $r$ \\
		Train with Adam Optimizer and Sparse Categorical Loss Function \\
		Save $M$ \ \ \& \ \ $M=300$ in this study. \\
		Predict($d_{\textit{training}}$) \\
		Save Predictions[l,p] \& $l=$labels, $p=$probability distributions
	}
	Use $[l,p]$, $d_{\textit{training}}$ and $d_{\textit{testing}}$ as inputs to SOCP model \ \ \& \ \ Please refer to the Equation (\ref{convexq2})
	Solve SOCP and get $w_{M,e}$ where $w=$weight vectors \\
	Prune $w$ according to $h$ on $d_{\textit{valid}}$\\
	Apply Voting \\
	Get P.
	\caption{Pruning the ensemble of CNNs with SOCP}\label{STGY}
\end{algorithm}
In the Equation (\ref{convexq2}),$\alpha$ and $\lambda$ values are hyperparameters that need to be tuned by using cross-validation. The decided range for these parameters for each dataset was $\alpha$ $=$ $(0,1; 0,2; 0,3; 0,4; 0,5)$ and for $\lambda$ $=$ $(0,1; 0,3; 0,5; 0,7; 0,9)$ where the best values are determined by cross-validation. The solution of the SOCP model given by  Equation (\ref{convexq2}) produces the weight for each model. After obtaining the sparse weights, the CNN models corresponding to non zero weights are chosen within the ensemble which lead to the subensemble. The final decision of the classification was performed by voting among the models in the subensemble. Several thresholds were experimented during validation which leads to different pruning extents (number of remaining models) and it was shown that favorable results were maintained as indicated in Table (\ref{prunedfull}). Threshold values were decided on the training set for weights approaching zero in sparse solutions of the Equation (\ref{convexq2}). The predictions of the selected models are then combined through a Voting aggregation to obtain the final predictions that are used to get test scores. The sparse solution of the Equation (\ref{convexq2}), the weight values closest to zero were taken and voting is performed by choosing CNN models in the corresponding ensemble. The voting corresponds to the weighted sum equation which is given in Equation (\ref{eq1}) since the target values are class labels in the classification problem. The weighted sum can be taken directly in a regression problem. 

 In Table \ref{prunedfull}, the prediction values (test accuracy) of the model on the test set; number of models and threshold values on pruned and full ensemble models with voting are given. Table \ref{bestavgmin} shows maximum accuracy estimation (Max. Accuracy score on test set), average accuracy estimation (Avg. Accuracy score on test set) and minimum accuracy estimation of pruned ensemble models and full ensemble models without using the voting method. The resulting $56$ pruned models for CIFAR-10, $98$ pruned models for CIFAR-100 and $170$ pruned models for MNIST performed close to performance of full models which is $300$ and it can be seen that pruned models are more advantageous in terms of time and complexity than unpruned models.

\begin{table}[]
	\centering
	\caption{Performance Results of Full and Pruned Ensemble}
	\label{prunedfull}
	\resizebox{8cm}{3.5cm}{
		\begin{tabular}{|l|cc}
			\hline
			\multicolumn{1}{|c|}{\textbf{ }} &	\multicolumn{1}{c|}{\textbf{Full Ensemble}} &	\multicolumn{1}{c|}{\textbf{Pruned Ensemble}} \\ \hline
			\textit{\textbf{CIFAR-10}}	& \multicolumn{2}{c|}{}  										\\ \hline
			Test Accuracy      			& \multicolumn{1}{c|}{90.33\%} 	& \multicolumn{1}{c|}{90.12\%}  \\ \hline
			Number of Models   			& \multicolumn{1}{c|}{300} 		& \multicolumn{1}{c|}{56} 		\\ \hline
			Threshold        			& \multicolumn{1}{c|}{N/A} 		& \multicolumn{1}{c|}{0.131} 	\\ \hline
			\textit{\textbf{CIFAR-100}}	& \multicolumn{2}{c|}{}  										\\ \hline
			Test Accuracy     		 	& \multicolumn{1}{c|}{59.54\%} 	& \multicolumn{1}{c|}{59.28\%}  \\ \hline
			Number of Models 		  	& \multicolumn{1}{c|}{300} 		& \multicolumn{1}{c|}{98} 		\\ \hline
			Threshold        			& \multicolumn{1}{c|}{N/A} 		& \multicolumn{1}{c|}{0.188} 	\\ \hline
			\textit{\textbf{MNIST}}		& \multicolumn{2}{c|}{}  										\\ \hline
			Test Accuracy    		  	& \multicolumn{1}{c|}{99.66\%} 	& \multicolumn{1}{c|}{99.64\%}  \\ \hline
			Number of Models 		  	& \multicolumn{1}{c|}{300} 		& \multicolumn{1}{c|}{170} 		\\ \hline
			Threshold        			& \multicolumn{1}{c|}{N/A} 		& \multicolumn{1}{c|}{0.18} 	\\ \hline
		\end{tabular}%
		}
\end{table}

\begin{table}[]
	\centering
	\caption{Best, average and worst performance results for CIFAR-10, CIFAR-100 and MNIST}
	\label{bestavgmin}
	\resizebox{8cm}{4.5cm}{
		\begin{tabular}{|l|cc}
			\hline
			\multicolumn{1}{|c|}{\textbf{ }} &	\multicolumn{1}{c|}{\textbf{Full Ensemble}} &	\multicolumn{1}{c|}{\textbf{Pruned Ensemble}} \\ \hline
			\textit{\textbf{CIFAR-10}}		& \multicolumn{2}{c|}{}  										\\ \hline
			Max. Accuracy Score     		& \multicolumn{1}{c|}{86.11\%} 	& \multicolumn{1}{c|}{84.44\%}  \\ \hline
			Avg. Accuracy Score   			& \multicolumn{1}{c|}{81.28\%} 	& \multicolumn{1}{c|}{79.10\%} 	\\ \hline
			Min. Accuracy Score      		& \multicolumn{1}{c|}{69.23\%} 	& \multicolumn{1}{c|}{69.23\%} 	\\ \hline
			Number of Models   				& \multicolumn{1}{c|}{300} 		& \multicolumn{1}{c|}{56} 		\\ \hline
			\textit{\textbf{CIFAR-100}}		& \multicolumn{2}{c|}{}  										\\ \hline
			Max. Accuracy Score     		& \multicolumn{1}{c|}{49.45\%} 	& \multicolumn{1}{c|}{49.45\%}  \\ \hline
			Avg. Accuracy Score   			& \multicolumn{1}{c|}{33.36\%} 	& \multicolumn{1}{c|}{34.08\%} 	\\ \hline
			Min. Accuracy Score      		& \multicolumn{1}{c|}{11.52\%} 	& \multicolumn{1}{c|}{11.52\%} 	\\ \hline
			Number of Models   				& \multicolumn{1}{c|}{300} 		& \multicolumn{1}{c|}{98} 		\\ \hline
			\textit{\textbf{MNIST}}			& \multicolumn{2}{c|}{}  										\\ \hline
			Max. Accuracy Score     		& \multicolumn{1}{c|}{99.64\%} 	& \multicolumn{1}{c|}{99.63\%}  \\ \hline
			Avg. Accuracy Score   			& \multicolumn{1}{c|}{99.24\%} 	& \multicolumn{1}{c|}{99.23\%} 	\\ \hline
			Min. Accuracy Score      		& \multicolumn{1}{c|}{95.22\%} 	& \multicolumn{1}{c|}{97.86\%} 	\\ \hline
			Number of Models   				& \multicolumn{1}{c|}{300} 		& \multicolumn{1}{c|}{170} 		\\ \hline
		\end{tabular}%
	}	
\end{table}
\section{Conclusions and Future Work} \label{conclusion}
This study aims to develop a mathematical model which prunes the ensemble of CNNs which are in response to the need in the literature with sparse SOCP. The strength of this paper is improving the performance and decreasing the complexity of the ensemble of CNNs with pruning. An important part to be considered in modeling the pruning algorithm is the accuracy performance of the networks at different depths which will be randomly generated and the richness of diversity among each other. However, there is a trade off between these two terms, in other words, the diversity is compromised as the percentage of accuracy increases. The mathematical model which is developed in this study is the pruning algorithm that keeps this trade off at the optimum level. The proposed pruning algorithm is a generic algorithm by being independent of the domain of the data set. Success rates are compared of pruned and full ensemble models by applying them in different areas of data sets. It has been observed that the difference between experiments is around $2\%$ and it can be said that the system is robust. Since the number of models are reduced significantly, similar or better accuracy scores are obtained in all data sets.

Recent studies in the literature \cite{Bruno22,Koles20,Sun17} leads to the conclusion that the use of pre-trained models and transfer learning increases the performance of the model and ensembling techniques outperform other state of the art models. For this reason, it is planned to test the SOCP model on an ensemble that will be created by using the models in the literature instead of the ensemble we randomly generated as a future work.
\bibliography{pruningdnn-template}
\newpage
\appendix
\section{Convex Form of Shannon Entropy} \label{shan}
Shannon entropy is defined for a given discrete probability distribution; measures how much information is required on average to identify random samples from this distribution. Shannon Entropy of a function A, which is considered as entropy here, is defined as follows \cite{Shannon48}:
\begin{equation}
	\centering
	H(A) = - \sum_{i=1}^{n} p_i \log p_i.
	\label{shannon}
\end{equation}
The $p$ values given by Equation (\ref{shannon}) refer to the probability measures such that if $p(a_i)=1/n$ is obtained for all "$i$" values, then $H(A) \leq \log_n$. The function $y=\log x$ is concave because the second derivative is less than zero. Since the $-\log$ term given in the equation is convex, Jensen's inequality is given as follows;
\begin{equation}
	-H(A) = - \log \Bigg(\sum_{i=1}^{n} \frac{1}{p_i} p_i \Bigg) \leq \sum_{i=1}^{n} \Bigg(- \log \frac{1}{p_i} \Bigg) p_i.
	\label{Jensen}
\end{equation}
The Jensen inequality expressed by Equation (\ref{Jensen}) holds unless $p_1=\ldots=p_n$ and $H$ is strictly concave with respect to $p$. Starting from the Jensen inequality, it can be said that if $H$ is concave, then $-H$ is strictly convex. If it considers it according to a function; let $F[p]=-H(p)$; its functional derivative is $\frac{\delta F[p]}{\delta p(z)}(p(z))=\log p(z)+1$ \cite{Baez11,Nish20}.

\section{Relaxations for Convex Quadratic Functions} \label{quad}
Convex optimization problems that minimize the cartesian product of quadratic (Lorentz) cones on the intersection of a linear function and affine linear manifold are defined as second-order cone programming (SOCP) problems. Linear problems, convex quadratic programs, and second-order constrained convex quadratic programs can all be defined as SOCP problems \cite{Alizadeh03}. The problem given by Equation (\ref{socp1}) is defined as a general quadratic conic programming \cite{Lobo98}:

\begin{equation}
	\begin{aligned}
		\min \quad & f^{T}x\\
		\textrm{s.t.} \quad & \| A_{i}x=b_i  \| \leq c^{T}_i x +d_i, \ \ \ \ i=1,\ldots,N,\\  
	\end{aligned}
	\label{socp1}
\end{equation}

where $x\in R^{n}$ is an optimization variable and parameters of the problem are as follows; $f\in R^{n}, A_i \in R^{(n_i-1)n}, b_i \in R^{n_i-1}, c_i \in R^n$ and $d_i \in R$. The norm seen in the constraints of the Equation (\ref{socp1}) is the standard Euclidean norm; $\|u\|=(u^{T}u)^{1/2}$. The constraint $\| A_{i}x=b_i  \| \leq c^{T}_i x +d_i$ is the second order conic constraint with dimension $n_i$. The $k$ dimension of a unit second-order convex cone is defined as:

\begin{equation*}
	\centering
	\ell_k= \Bigg\{\left[\begin{matrix} u\\t\\\end{matrix}\right]\ |\ u\in\ R^{k-1},\ t\in\ R,\ u \leq t \Bigg\},
\end{equation*}
when $k=1$, the unit second degree conic turns into the following:

\begin{equation*}
	\ell_1={t\ |\ t\in\ R,\ 0\leq t}. 
\end{equation*}

The set of points satisfying a constraint of the quadratic conic problem is the inverse image of the unit quadratic cone according to affine mapping;

\begin{equation*}
	\| A_{i}x=b_i  \| \leq c^{T}_i x +d_i \Leftrightarrow A_i c^{T}_i x +b_i d_i \in \ell_{n_{i}},
\end{equation*}

therefore, it is convex \cite{Lobo98}.

In quadratic optimization problems with convex structure, each local minimizer is a global minimizer and an approximation to global minimizer can be calculated. Such problems can be converted to second-order conic problems and solved by applying primal-dual interior point methods \cite{Kim01}. Generally, convex quadratic function can be illustrated with:

\begin{equation*}
	f(x)= x^{T}Qx+c^{T}x,
\end{equation*}

if and only if $Q$ is positive semi-definite (PSD) and $Q$ is PSD only when $x\geq 0$. If a quadratic term $t$ is introduced, then the equation can be rewritten as follows:

\begin{equation*}
	\begin{aligned}
		\min \quad & t+c^{T}x\\
		\textrm{s.t.} \quad & Ax=b,\\
		& t \geq x^{T}Qx .   \\  
	\end{aligned}
\end{equation*}

It is observed that the last added constraint can be expressed using a rotated quadratic cone. If expressed mathematically; the inequality $t\geq x^TQx$ is satisfied in the case of $\left(Q^\frac{1}{2}x,t,1\right)\in Q_{rot}^n$.

The convex quadratic cone programming problem with limited constraints is a nonlinear programming problem that can be viewed as a trust region subproblem in the trust-region method \cite{Zhang12,Kato07}. Especially strong convex quadratic problems have polyhedral feasible regions like linear problems and these type of problems can be expressed as a second order cone problem:
\begin{equation}
	\label{socp5}
	\begin{aligned}
		\min \quad & q (x)=x^TQx+a^Tx+\beta\\
		\textrm{s.t.} \quad & Ax=b,\\ 
		& x\geq0,  \\  
	\end{aligned}
\end{equation}
where $Q$ in Equation (\ref{socp5}) is the symmetric positive definite matrix; $Q>0$,$Q=Q^T$. The objective is $q(x)=\|\bar{u}\|^{2}+\beta-\frac{1}{4}a^{T}Q^{-1}a$, and $\bar{u}=Q^{\frac{1}{2}}x + \frac{1}{2}Q^{-\frac{1}{2}}a$. Therefore, the problem can be written in SOCP form as in Equation (\ref{socpforg};
\begin{equation}
	\label{socpforg}
	\begin{aligned}
		\min \quad & u_0\\
		\textrm{s.t.} \quad & Q^{1/2}x-\bar{u}=\frac{1}{2}Q^{-1/2}a,\\ 
		& Ax=b,   \\
		& 	x\geq0. 
	\end{aligned}
\end{equation}
Although it gives the same optimal result in both problems, the change is up to $\beta-\frac{1}{4}a^TQ^{-1}a$ in the main problem \cite{Alizadeh03}. Convex quadratic problems are expressed as minimization of a linear function with convex quadratic constraints;
\begin{equation}
	\label{convexq}
	\begin{aligned}
		\min \quad & c^{T}x\\
		\textrm{s.t.} \quad & q_i(x)=x^TB_i^TB_ix+a_i^Tx+\beta i\leq0,\\ 
		& i=1,\ldots,m.   \\
	\end{aligned}
\end{equation}
In Equation (\ref{convexq}) above, $B_i\in R^{k_ixn}$ and rank $k_i$, $i=1,\ldots,m$. The constraint $q(x)=x^TB^TBx+a^Tx+\beta\leq0$ means the same thing as $(u_0;\bar{u})\geq Q_0$ given as the second-order conic constraint:
\begin{equation*}
	\bar{u}=\ \left(\begin{matrix}Bx\\\frac{a^Tx+\beta+1}{2}\\\end{matrix}\right),
\end{equation*}
\begin{equation*}
	u_0=\frac{1-a^Tx-\beta}{2}.
\end{equation*}
\end{document}